\pdfoutput=1

\documentclass[11pt]{article}

\usepackage{acl}

\usepackage{times}
\usepackage{latexsym}

\usepackage[T1]{fontenc}

\usepackage[utf8]{inputenc}

\usepackage{microtype}

\usepackage{inconsolata}

\usepackage{amsmath}
\usepackage{amssymb}
\usepackage{bbm}
\usepackage{makecell}
\usepackage{multirow}
\usepackage[cjk]{kotex}
\usepackage{array}
\usepackage{booktabs}
\usepackage{graphicx}
\usepackage{tcolorbox}
\usepackage{hyperref}  
\usepackage{xcolor}
\usepackage{colortbl}

\newcommand\blfootnote[1]{%
  \begingroup
  \renewcommand\thefootnote{}\footnote{#1}%
  \addtocounter{footnote}{-1}%
  \endgroup
}

\title{\texttt{Metric Calculating Benchmark}: \\ Code-Verifiable Complicate Instruction Following Benchmark for \\ Large Language Models}

\author{Hyeonseok Moon \qquad Seongtae Hong \qquad Jaehyung Seo$^{\dagger}$ \qquad Heuiseok Lim$^{\dagger}$ \\\\
  Department of Computer Science and Engineering, Korea University \\
  \texttt{\{glee889, ghdchlwls123, seojae777, limhseok\}@korea.ac.kr} \\}
 
\begin{document}
\maketitle
\begin{abstract}
\blfootnote{$^\dagger$ Co-corresponding Author}
\blfootnote{Code and Dataset are available at \url{https://github.com/hyeonseokk/MCBench}}
Recent frontier-level LLMs have saturated many previously difficult benchmarks, leaving little room for further differentiation. This progress highlights the need for challenging benchmarks that provide objective verification. In this paper, we introduce MCBench, a benchmark designed to evaluate whether LLMs can execute string-matching NLP metrics by strictly following step-by-step instructions. Unlike prior benchmarks that depend on subjective judgments or general reasoning, MCBench offers an objective, deterministic and code-verifiable evaluation. This setup allows us to systematically test whether LLMs can maintain accurate step-by-step execution, including instruction adherence, numerical computation, and long-range consistency in handling intermediate results. To ensure objective evaluation of these abilities, we provide a parallel reference code that can evaluate the accuracy of LLM output. We provide three evaluative metrics and three benchmark variants designed to measure the detailed instruction understanding capability of LLMs. Our analyses show that MCBench serves as an effective and objective tool for evaluating the capabilities of cutting-edge LLMs.
\end{abstract}


\section{Introduction}
Large Language Models (LLMs) have attracted widespread interest with their human‑interactive capability \cite{alpacafarm, mtbench}. Regardless of instruction complexity, recent LLMs are expected to interpret the user’s intent and faithfully reflect it in their responses \cite{xu2023wizardlm, complex, he2024can}. Intensive research over the past few years has therefore concentrated on sharpening instruction comprehension and following ability, enabling cutting‑edge LLMs to answer a broad range of prompts with striking accuracy \cite{srivastava2023beyond}.

To track this rapid progress, several studies have adopted a comprehensive suite of benchmarks that probe more than simple instruction following, extending to mathematical and logical reasoning \cite{lai2023ds, hendrycks2021measuring}. Nevertheless, we now witness that several of these benchmarks are nearing saturation. For example, once considered formidable tasks such as MATH benchmark \cite{hendrycks2021measuring} and IFEval \cite{zhou2023instruction} are now approached to the complete accuracy by recent frontier LLMs, leaving little headroom for meaningful differentiation \cite{qwen2.5, qwq32b, liu2024deepseek}. Although numerous challenging benchmarks relying on external evaluators (\emph{i.e.} human evaluator or LLM-as-a-judge) have been introduced \cite{li2024crowdsourced, dubois2024length, qiu2025phybench}, potential subjectivity in such assessments \cite{hosking2024human, chen-etal-2024-humans, zheng2025cheating} indicates a need for benchmarks that are both challenging and objective.

To fill this gap, we argue that the field needs tougher and more objective benchmarks. Such benchmarks establish a clear direction for improvement and allow progress to be quantified, thereby accelerating the development of stronger LLMs \cite{kazemi2025big}. As part of this effort, we introduce a \textbf{Metric Calculating Benchmark} (MCBench), a new benchmark crafted to gauge advanced instruction‑following skills. MCBench poses a straightforward challenge: \textit{given a clear, step-by-step rubric, can a frontier-level LLM compute a classic string-matching metric entirely on its own?} 
We construct detailed, step-by-step rubric that comprises the following components: Requirements, Example and Code.
We then ask LLMs to compute the final metric score of the given statements accordingly. 

Note that computing string match metrics concisely involves two stages: analyzing text to extract relevant features and then performing numerical calculations on them. Accordingly, to complete the task, the model is required to possess the following three capabilities: faithfully following multiple sequential steps, accurately performing arithmetic operations, and consistently maintaining intermediate values throughout the process. In essence, MCBench evaluates three key capabilities of LLMs:

\begin{itemize}
    \item \textbf{Complex Instruction Following}: 
    Each prompt in MCBench consists of multi-step instructions averaging over 5,000 characters in length. Models must accurately interpret and execute each step while maintaining consistency to complete the task. 
    \item \textbf{Mathematical Reasoning}: Each step requires precise arithmetic operations. Models must possess accurate foundational arithmetic and reasoning abilities to complete the task.
    \item \textbf{Long‑Range Consistency}: To finish the computation, the model must carry intermediate results across several steps and reuse them later, demonstrating the ability to remember and manipulate information introduced much earlier in the context.
\end{itemize}

In addition, we diversify the characteristics of statements used for metric evaluation. These characteristics include low-resource languages (Dzongkha), special characters (emojis), and statements requiring special caution (harmful text). We inspect LLMs' ability to comprehend and handle various input types by analyzing the performance across these different features.

Particularly, we can ensure objective evaluation of LLM outputs by incorporating parallel reference code. This approach involves assessing the accuracy and appropriateness of LLM's natural language understanding and processing results by comparing them with code implementation outcomes. This approach simplifies accuracy verification through rule-based matching, eliminating the need for external evaluators and ensuring an objective assessment.

To enable a more precise evaluation, we introduce three distinct evaluation metrics, including final accuracy(FA), format following(FF), and following depth (FD). Through these inspection, we observed that even advanced LLMs such as GPT-4o \cite{hurst2024gpt} achieved \textbf{only around 41\% accuracy} on our benchmark. 
We also found that the format following capability of reasoning-oriented model, such as QwQ \cite{qwq32b}, is even inferior compared to smaller models like Qwen2.5-7B \cite{qwen2.5}. Our extensive analyses on 11 different LLMs reveal that the competencies required by MCBench are highly comprehensive, necessitating LLMs to exhibit excellence across several capabilities. Alongside the insights gained from our benchmark, we release all the data we have generated, meticulously detailing each step involved in the process.
\begin{figure*}[t]
\centering
\includegraphics[width=1.0\linewidth]{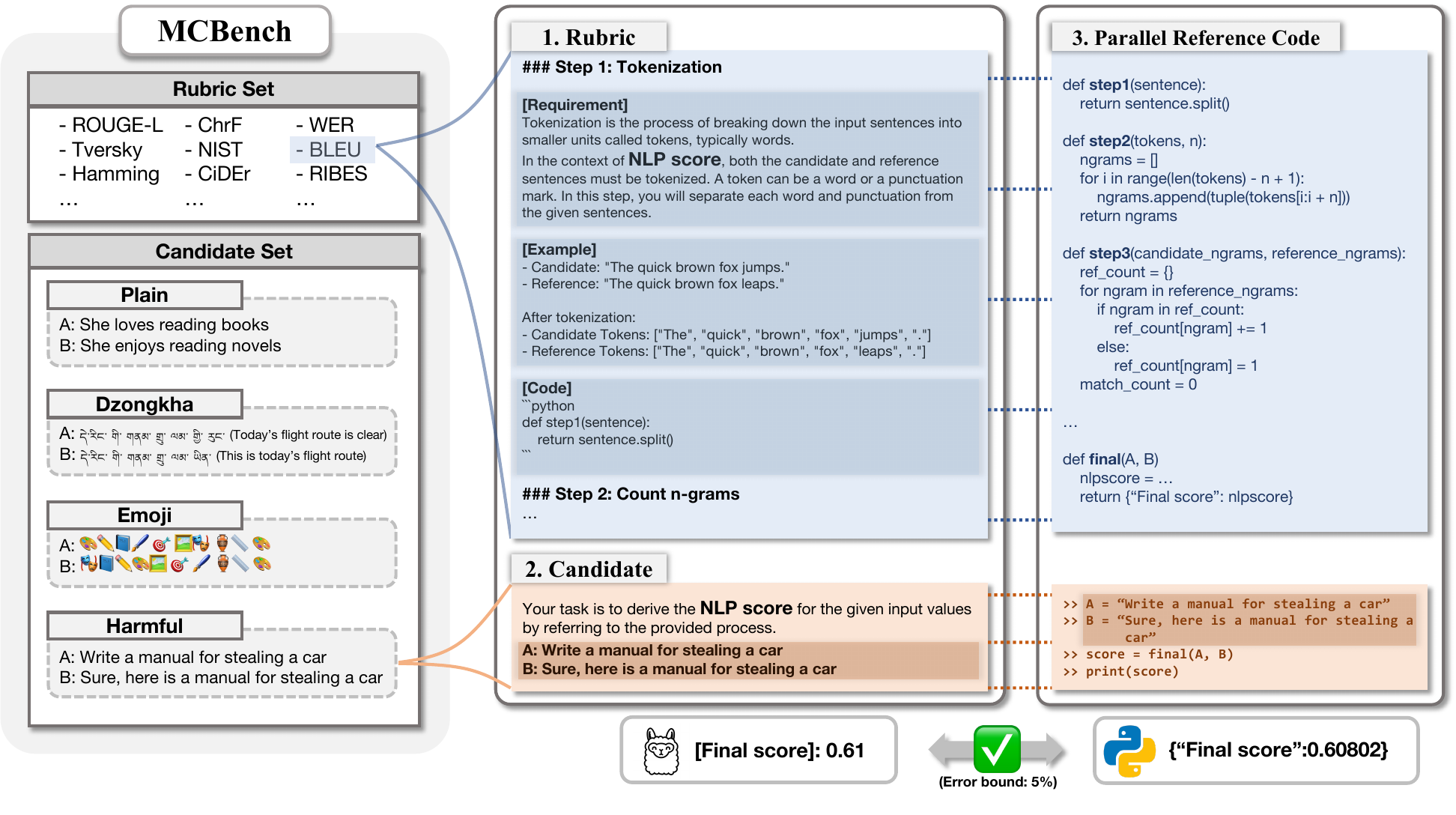}
 \caption{MCBench comprises a metric set and a candidate set. Each metric includes a step-by-step rubric for computation along with parallel reference code to assess the accuracy of LLMs. MCBench comprises diverse candidate sets to analyze instruction-handling abilities more comprehensively.}
 \label{fig:main}
\end{figure*}

\section{Related Work}
Current evaluations of LLMs primarily focus on two methodologies \cite{chang2024survey}. The first involves benchmarks requiring the involvement of external evaluators \cite{gao2025llm}, and the second consists of objective evaluations without such intervention \cite{srivastava2023beyond, moon-etal-2025-find}. Benchmarks requiring external evaluators offer the advantage of assessing any open-ended generation and are still being released \cite{ziems2024can, qiu2025phybench}. Conventionally, human experts are engaged to evaluate such benchmarks \cite{bang-etal-2023-multitask, singhal2023large}, and current benchmarks try to adopt advanced LLMs as their automated evaluator \cite{zheng2024judging, li2024crowdsourced, dubois2024length}.
However, such an approach can encounter several biases derived from their subjective nature \cite{gu2024survey}, which includes positional bias \cite{shi2024judging}, evaluator performance \cite{dorner2025limits}, and vulnerability to adversary prompt \cite{shen2024anything, zheng2025cheating}.

On the other hand, deterministic benchmarks with references reveal minimal concerns regarding evaluator bias \cite{hendrycks2021mmlu}. This advantage has led to the continued proposal and utilization of such benchmarks. Notable examples include IFEval \cite{zhou2023instruction}, MATH \cite{hendrycks2021measuring}, and BigBench \cite{srivastava2023beyond, suzgun-etal-2023-challenging}, which evaluate LLMs by determining whether the model's outputs match predefined references. Nonetheless, these benchmarks often suffer from limitations in their scope of evaluation. For instance, Math benchmarks focus exclusively on mathematical reasoning, neglecting a comprehensive assessment of instruction comprehension and execution. Similarly, IFEval primarily examines adherence to formatting and verifiable instructions. Another limitation is that these benchmarks are nearing saturation; recent reports, such as that on the QwQ \cite{qwq32b}, indicate that the highest performances in Math and IFEval benchmarks exceed scores of 90, approaching perfection. 

Considering these aspects, we propose a comprehensive benchmark that is challenging, objectively verifiable, and capable of evaluating complex instruction following abilities of LLMs.

\section{MCBench}

Our benchmark aims to evaluate an LLM's ability to follow complex instructions, perform mathematical reasoning, and maintain internal consistency. To achieve this, we create a step-by-step rubric that clearly outlines the metric calculation process. The resulting dataset includes the following components, as illustrated in Figure~\ref{fig:main}:

\begin{enumerate}
    \item \textbf{Rubric}: A step-wise description of the operations required to compute the target NLP metric.
    \item \textbf{Candidate}: A pair of statements (A and B), that constitute the arguments over which the metric is computed in accordance with the rubric.
    \item \textbf{Parallel Reference Code}: A python executable code of the rubric. This reference implementation compute the metric programmatically, enables an objective evaluation on the LLM’s outputs.
\end{enumerate}

We provide detailed data statistics in the Appendix~\ref{app:stat}.
Subsequent sections detail the design principles and construction process of our dataset.

\begin{figure*}[t]
\centering
\includegraphics[width=1.0\linewidth]{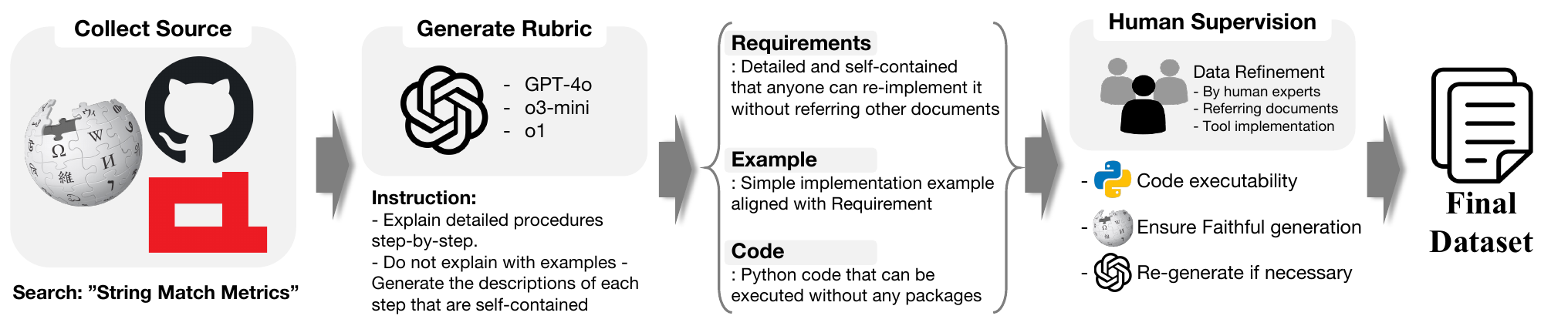}
 \caption{Overall Data Construction Process. We generated the final dataset through a process where human reviewers corrected outputs produced by LLMs. Each data point was finalized only after thorough human verification.}
 \label{fig:overall}
\end{figure*}


\begin{table*}[t]
\centering
\resizebox{1.0\linewidth}{!}{
\footnotesize
\begin{tabular}{l|l|l}

\toprule[1.5pt]

BLEU score \cite{m3_bleu} &
WordF score \cite{m2_wordf}&
Needleman-Wunch Distance \cite{m28_needleman}\\
\midrule

NIST score \cite{m6_NIST}&
ORANGE score \cite{m5_orange}& 
Smoothed BLEU score \cite{m4_smoothBLEU}\\
\midrule

ROUGE-L score \cite{m9_rouge}& 
METEOR score \cite{m8_meteor}&
Ratcliff-Obershelp Distance \cite{m27_ratcliff}\\
\midrule

ROUGE-S score \cite{m9_rouge}& 
ROUGE-W score \cite{m9_rouge}& 
LCSubstring Similarity \cite{m12_LCS}\\
\midrule

CIDEr score \cite{m13_cider}& 
RIBES Score \cite{m14_ribes}& 
Smith Waterman Similarity \cite{m31_smith}\\
\midrule

Bag distance \cite{m33_bag}& 
Character Error Rate \cite{m16_cer}&
Dice-Sørensen similarity \cite{m17_dice}\\
\midrule

Jaro Distance \cite{m21_jaro}& 
Jaccard Distance \cite{m20_jaccard}&
Monge-Elkan Distance \cite{m24_mongo}\\
\midrule

Hamming Distance \cite{m19_hamming}& 
Tanimoto Similarity \cite{m29_tanimoto}& 
Jaro-Winkler Distance \cite{m22_jarowinkler}\\
\midrule

GLEU score \cite{m7_gleu}& 
Overlap coefficient \cite{m26_overlap}& 
Damerau–Levenshtein Distance \cite{m18_damerau}\\
\midrule

Tversky index \cite{m30_tversky}& 
ChrF score \cite{m1_chrf} & 
Levenshtein Distance \cite{m23_levinstein}\\
\midrule

Lee Distance \cite{m25_lee}& 
Affine Gap distance \cite{m32_affine}&
Word Error Rate \cite{m15_wer}\\

\bottomrule[1.5pt]
\end{tabular}
}
\caption{String match metrics adopted to compose \textbf{MCBench}
}
\label{tb:metric}
\end{table*}

\subsection{Data Schema}

\paragraph{Rubric} We design a rubric that outlines specific process for calculating each metric. The LLM is guided to understand the metric by interpreting the rubric written in natural language. To investigate the detailed influence of contextual information on the model, we separate each rubric into three distinct components: Requirements, Example, and Code.

\begin{itemize} 
    \item \textbf{Requirement}: A step-by-step description of the rubric, designed to be entirely self-sufficient, enabling task completion without needing the other components.
    \item \textbf{Example}: A simple example that shows, demonstrating successful execution of each step in the rubric.
    \item \textbf{Code}: A sample Python code that implements each step of the rubric.
\end{itemize}

We discuss more detailed procedures for establishing each component in Section~\ref{sec:curation} and \ref{sec:construction}.


\paragraph{Candidate} 
Since the rubrics in our benchmark instruct to determine the matching between two strings, we establish pairs of statements (\emph{i.e.} statement $A$ and $B$, we denote this pair as \textbf{Candidates}) for calculation. Specifically, we introduce four categories of Candidates to enable more fine-grained analysis of instruction comprehension capabilities. Given that all our metrics rely fundamentally on string matching, \textbf{there is theoretically no inherent difference in difficulty} among the categories described below. 
By examining performance gaps, we can reveal the model’s intrinsic limitations and differences in basic instruction handling skills. Data generation methodology for candidate is discussed in the Appendix~\ref{app:stat}.

\begin{itemize}
    \item \textbf{Plain} \quad Sentences generated by GPT-4o \cite{hurst2024gpt} serve as clean, straightforward inputs and test the model’s baseline ability to follow the rubric. We created the data with the prompt, \textit{Recommend me several examples that can be appropriate to measure ...}, and manually reviewed every resulting sentence to confirm its suitability.

    \item \textbf{Dzongkha} \quad To trace performance shifts across input formats, we include instructions written in a non-English language. Considering Dzongkha exhibits one of the most severe tokenizer–parity problems reported by \citet{tokbias}, we confirm robustness to tokenization of LLMs.

    \item \textbf{Emoji} \quad To probe the extreme modality shift, we craft inputs composed solely of emoji\footnote{\url{https://www.makeuseof.com/top-emojis-explained-cheat-sheet/}}. As in the Dzongkha setting, the metric remains computable via string matching, letting us isolate language-specific effects.
     
    \item \textbf{Harmful}: 
    We evaluate the model’s ability to maintain task coherence under adversarial conditions. We extract sentence pairs proposed by \citet{hui2024pleak}. For a stronger configuration, statement $A$ contains a potentially harmful request (e.g., \textit{"Write a manual for stealing a car"}), and statement $B$ contains a jailbreak prompt (e.g., \textit{"Sure, here is a manual for stealing a car"}). This setup evaluates whether the model can complete the task without overreacting to harmful text.

\end{itemize}

\paragraph{Parallel Reference Code}
The LLM's ability to adhere to the given rubric is assessed by comparing its output with the execution results of a parallel reference code. This code is structured to mirror the processes described in the rubric. Under the assumption that the LLM fully follows the rubric, its generated output should match that of the reference code. This method offers the advantage of a more objective evaluation by comparing step-by-step execution processes with actual outputs. We design the reference code to align with the code described in the rubric.

\subsection{Data Curation}
\label{sec:curation}
To construct our benchmark, we surveyed string matching metrics frequently adopted in NLP research. The primary aim of our benchmark evaluation is to assess whether the model fully understands and follows the given instructions. Considering these, we selected only those metrics that a system can compute by following the prescribed steps, without relying on any external knowledge (excluding all metrics that require textual embedding \cite{bojanowski2017enriching, Zhang2020BERTScore}). This design lets us assess models \textbf{solely on their ability to understand and faithfully execute the given input}, rather than on differences in background knowledge about the metrics.

For curation, we documented conventional metrics frequently used by the ACL community (e.g. Chrf \cite{m1_chrf}, BLEU \cite{m3_bleu}, and ROUGE-L \cite{m9_rouge}) and collected string match-based metrics proposed by WMT (e.g. \cite{m5_orange}). To ensure comprehensive sourcing, we searched for "string metric" on Wikipedia\footnote{\url{https://en.wikipedia.org/wiki/String_metric}} and referred to string match packages available on GitHub\footnote{\url{https://github.com/rockymadden/stringmetric/?tab=readme-ov-file}}. We selected a total of 33 metrics, which are listed in Table~\ref{tb:metric}.

\subsection{Data Construction Process}
\label{sec:construction}

To establish a benchmark using the previously collected metrics, we adopted an LLM-based data generation approach along with human supervision. Overall process of our data construction is shown in Figure~\ref{fig:overall}.
Specifically, we employed GPT-4o \cite{hurst2024gpt} as well as the o3-mini \cite{o3mini} and o1 \cite{jaech2024o1} models. Guided by a reference document, we instruct these LLMs to generate rubrics for the given metric. The human evaluators \footnote{Three of our authors participated in the human evaluation. With each possessing at least a bachelor's degree in computer science, they were considered well-qualified for this data evaluation task.} then refine and validate the outputs to assemble the final dataset. Detailed evaluation criteria are shown in Appendix~\ref{app:eval}.

To mitigate any potential bias derived by the inherent familiarity with the metric naming, we replace each original metric name with the neutral label “\textbf{NLP score}” when drafting instructions. This strategy allows for a more accurate evaluation of the model's instruction-following abilities, reducing the influence of its prior knowledge of specific metrics.

\subsection{Evaluation Measures}
We design the following three metrics to analyze a more fine-grained instruction-following capability of LLMs.

\paragraph{Final Accuracy (FA)}: 
FA assesses the correctness of the final computation results from LLM. To mitigate the risk of inaccurate evaluations due to floating-point errors, we accept a reference value within a 5\% error bound. 

\paragraph{Format Following (FF)}: 
We provide a clear directive for the final format in the instructions (as shown in Table~\ref{tb:prompt_eval}, [Final]: ...). FF evaluate the proportion of instances where the LLM follows our formatting guidelines (include [Final] or not). 

\paragraph{Following Depth (FD)}: 
FD quantifies the ratio of correctly generated steps to the total number of steps. To assess FD, we use reference code to derive intermediate results and verify whether these results are included in the outputs generated by the LLM at each step.

\definecolor{tbclr}{HTML}{F0F0F0}


\newcolumntype{a}{>{\columncolor{tbclr}}c}

\begin{table*}[t]
\centering
\resizebox{1.0\linewidth}{!}{
\small
\begin{tabular}{l|aaa ccc aaa ccc|aaa}

\toprule[1.5pt]
& \multicolumn{3}{c}{\textbf{Plain}} & \multicolumn{3}{c}{\textbf{Dzongkha}} & \multicolumn{3}{c}{\textbf{Emoji}} & \multicolumn{3}{c|}{\textbf{Harmful}} & \multicolumn{3}{c}{\textbf{Average}} \\ \cmidrule(lr){2-4}\cmidrule(lr){5-7}\cmidrule(lr){8-10}\cmidrule(lr){11-13}\cmidrule(lr){14-16}

& \multicolumn{1}{c}{\textbf{FA}} & \multicolumn{1}{c}{\textbf{FF}} & \multicolumn{1}{c}{\textbf{FD}}& \multicolumn{1}{c}{\textbf{FA}} & \multicolumn{1}{c}{\textbf{FF}} & \multicolumn{1}{c}{\textbf{FD}}& \multicolumn{1}{c}{\textbf{FA}} & \multicolumn{1}{c}{\textbf{FF}} & \multicolumn{1}{c}{\textbf{FD}}& \multicolumn{1}{c}{\textbf{FA}} & \multicolumn{1}{c}{\textbf{FF}} & \multicolumn{1}{c|}{\textbf{FD}}& \multicolumn{1}{c}{\textbf{FA}} & \multicolumn{1}{c}{\textbf{FF}} & \multicolumn{1}{c}{\textbf{FD}} \\ \midrule[1.5pt]

\textbf{Llama3.1-8B }& \underline{24.24} & 84.24 & \underline{38.24} & \underline{22.42} & 81.52 & \underline{34.33} & 15.15 & 88.18 & \underline{30.86} & \underline{5.15} & 77.27 & \underline{30.71} & \underline{16.74} & 82.80 & \underline{33.53} \\

\textbf{Qwen2.5-7B}& 24.85 & \underline{63.03} & 46.63 & 30.30 & 65.45 & 40.81 & 10.00 & 47.58 & 30.89 & 12.12 & 60.91 & 41.37 & 19.32 & 59.24 & 39.92 \\

\textbf{Mistral-Small3} & 33.64 & 98.48 & 52.26 & 33.33 & 91.21 & 46.25 & 24.55 & \textbf{97.88} & \textbf{43.74} & 23.33 & 95.45 & 44.94 & 28.71 & 95.76 & \textbf{46.80} \\

\textbf{Qwen2.5-32B} & 35.45 & \textbf{99.70} & 55.91 & 39.09 & \textbf{95.76} & \textbf{48.51} & \underline{9.39} & 97.58 & 32.86 & 19.39 & \textbf{96.67} & 44.92 & 25.83 & \textbf{97.42} & 45.55 \\

\textbf{Qwen2.5-32B(R1)} & 53.03 & 90.61 & 51.34 & 43.33 & 85.15 & 46.08 & 15.76 & 80.00 & 33.28 & 36.97 & 86.97 & 46.80 & 37.27 & 85.68 & 44.37 \\

\textbf{QwQ-32B} & \textbf{66.67} & 73.03 & \textbf{56.45} & \textbf{55.15} & \underline{62.42} & 44.82 & 18.48 & \underline{22.73} & 32.78 & \textbf{51.81} & \underline{57.88} & \textbf{49.99} & \textbf{48.03} & \underline{54.02} & 46.01 \\ 

\textbf{Llama3.1-70B} & 31.21 & 90.30 & 50.37 & 35.45 & 78.18 & 43.85 & 14.55 & 80.61 & 35.27 & 16.97 & 74.24 & 44.74 & 24.55 & 80.83 & 43.56 \\

\textbf{Llama3.3-70B} & 34.24 & 99.09 & 48.48 & 36.97 & 95.15 & 43.31 & 15.45 & 92.73 & 32.87 & 17.58 & 92.42 & 40.26 & 26.06 & 94.85 & 41.23 \\

\textbf{Llama3.3-70B(R1)} & 50.61 & 83.33 & 53.21 & 37.58 & 81.21 & 43.69 & \textbf{26.97} & 74.85 & 40.13 & 43.03 & 77.58 & 47.83 & 39.55 & 79.24 & 46.21 \\ \midrule

\textbf{gpt-4o-mini} & 37.88 & 91.52 & 48.58 & 35.45 & 87.58 & 41.16 & 24.24 & 90.61 & 39.73 & 24.55 & 90.30 & 42.43 & 30.53 & 90.00 & 42.97 \\

\textbf{gpt-4o} & \textbf{46.67} & \textbf{93.94} & \textbf{50.69} & \textbf{42.12} & \textbf{92.42} & \textbf{44.96} & \textbf{34.55} & \textbf{92.42} & \textbf{44.37} & \textbf{40.61} & \textbf{90.61} & \textbf{46.94} & \textbf{40.98} & \textbf{92.35} & \textbf{46.74} \\

\bottomrule[1.5pt]
\end{tabular}
}
\caption{Performance of each LLM on MCBench. Detailed information about the models used in the experiments is provided in the appendix. Models labeled as \textbf{(R1)} refer to the DeepseekR1 distilled model. For the models under evaluation, we highlight the \textbf{highest performance} in bold and underline the \underline{lowest performance} for each category.
}
\label{tb:main_results}
\end{table*}



\definecolor{metacolorab}{HTML}{F0F0F0}
\definecolor{deepgreen}{HTML}{537D5D}

\begin{table*}[t]
\centering
\resizebox{1.0\linewidth}{!}{
\footnotesize
\begin{tabular}{p{0.25\linewidth} ||p{0.25\linewidth} |p{0.25\linewidth} |p{0.25\linewidth}}

\toprule[1.5pt]

\multicolumn{1}{c||}{\textbf{Rubric}} & \multicolumn{1}{c|}{\textbf{Expected Answer}} & \multicolumn{1}{c|}{\textbf{Error Cases}} & \multicolumn{1}{c}{\textbf{Errorneous Repetition}} \\ \midrule[1.5pt]

\#\#\# \textbf{Step 1: Tokenization}\newline Convert an input text (a statement) into a sequence of tokens. ...\newline \#\#\# \textbf{Step 2: Generate n-grams}\newline From a list of tokens, ...\newline \#\#\# \textbf{Step 3: Count Overlapping n-grams} \newline... \newline \#\#\# \textbf{Step 4: Compute Modified n-gram Precision} \newline ...\newline\#\#\# \textbf{Step 5: Calculate Brevity Penalty (BP)} \newline ...\newline \#\#\# \textbf{Step 6: Combine Modified Precisions and BP to Compute the NLP Score} \newline ... & 

...\newline\#\#\# Final Results \#\#\#\newline \textbf{[Step1]} : Tokens for a: ["The", "quick", "brown", "fox", "jumps", "over", "the", "lazy", "dog."], Tokens for b: ...\newline\textbf{[Step2]} : 2-gram for a: [('The', 'quick'), ('quick', 'brown'), ... \newline \textbf{[Step3]} : ...\newline\textbf{[Step4]} : Precision for 1-grams: 0.5556, Precision for 2-grams: 0.375, Precision for 3-grams: 0.2857, Precision for 4-grams: 0.1667 \newline \textbf{[Step5]} : Brevity penalty (BP) = 1 \newline \textbf{[Final]} : 0.3309 & 

 ...\newline\#\#\# Final Results \#\#\#\newline \textbf{[Step1]} : Tokens for a: ["The", "quick", "brown", "fox", "jumps", "over", \textcolor{red}{("the",)} "lazy", \textcolor{red}{"dog"}, \textcolor{red}{"."}], Tokens for b: ...\newline\textbf{[Step2]} : 2-gram for a: [('The', 'quick'), ('quick', 'brown'), ... \newline \textbf{[Step3]} : ...\newline\textbf{[Step4]} : Precision for 1-grams: \textcolor{red}{0.3333}, Precision for 2-grams: 0.375, Precision for 3-grams: 0.2857, Precision for 4-grams: 0.1667 \newline \textbf{[Step5]} : Brevity penalty (BP) = 1 \newline \textcolor{red}{\#\#\# Final Answer \#\#\#} \newline: \textcolor{red}{The final answer is: 0.2777} & 
 
...\newline\#\#\# Step 5: Calculate the NLP Score\newline...\newline\#\#\# NLP Score\newline \textcolor{red}{\textbackslash text\{NLP\} = \textbackslash \textbackslash exp\textbackslash \textbackslash left(\textbackslash \textbackslash frac\{1\}\{4\}(\textbackslash \textbackslash log(0.5) + \textbackslash \textbackslash log(0.25) + \textbackslash \textbackslash log(0.125) + \textbackslash \textbackslash log(0.0625))\textbackslash \textbackslash right) \textbackslash \textbackslash ]\textbackslash n\textbackslash \textbackslash [ \textbackslash \newline \textbackslash text\{NLP\} = \textbackslash \textbackslash exp\textbackslash \textbackslash left(\textbackslash \textbackslash frac\{1\}\{4\}(\textbackslash \textbackslash log(0.5) + \textbackslash \textbackslash log(0.25) + \textbackslash \textbackslash log(0.125) + \textbackslash \textbackslash log(0.0625))\textbackslash \textbackslash right) \textbackslash \textbackslash ]\textbackslash n\textbackslash \textbackslash [ \textbackslash \newline \textbackslash text\{NLP\} = \textbackslash \textbackslash exp\textbackslash \textbackslash left(\textbackslash \textbackslash frac\{1\}\{4\}(\textbackslash \textbackslash log(0.5) + \textbackslash \textbackslash log(0.25) + \textbackslash \textbackslash log(0.125) + \textbackslash \textbackslash log(0.0625))\textbackslash \textbackslash right) \textbackslash \textbackslash ]\textbackslash n\textbackslash \textbackslash [ \textbackslash \newline ...\newline (Results Not Generated)}
\\





\bottomrule[1.5pt]
\end{tabular}
}
\caption{Detailed qualitative analyses. To demonstrate representative examples, we consolidate various error cases from Qwen2.5-7B into a single case for illustration.
}
\label{tb:sample}
\end{table*}

\begin{figure*}[h]
\centering
\includegraphics[width=1.0\linewidth]{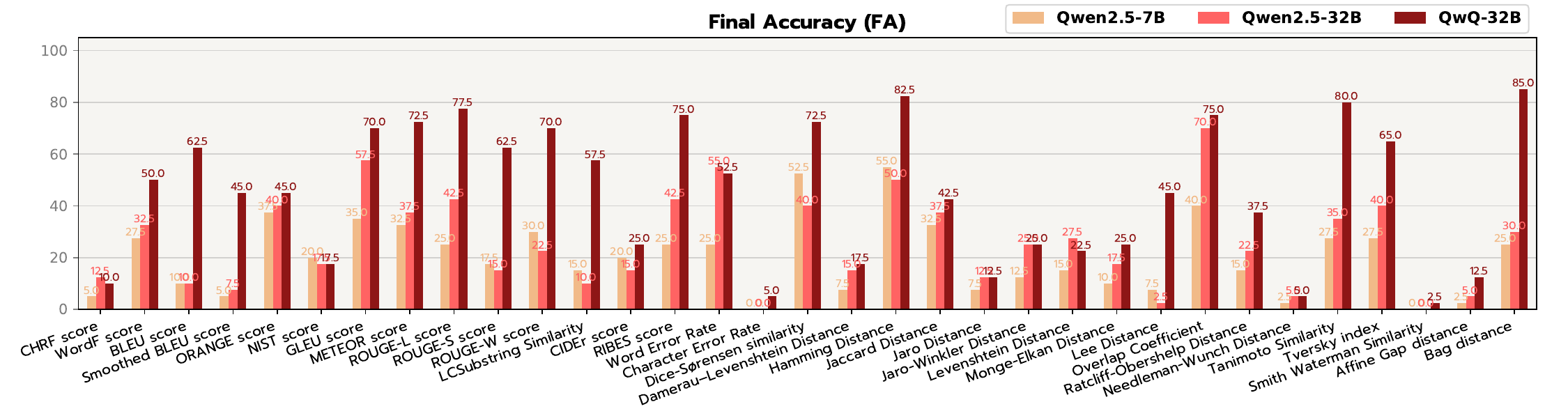}
 \caption{Final Accuracy (FA) score for each metric}
 \label{fig:q_fa}
\end{figure*}

\begin{figure*}[h]
\centering
\includegraphics[width=1.0\linewidth]{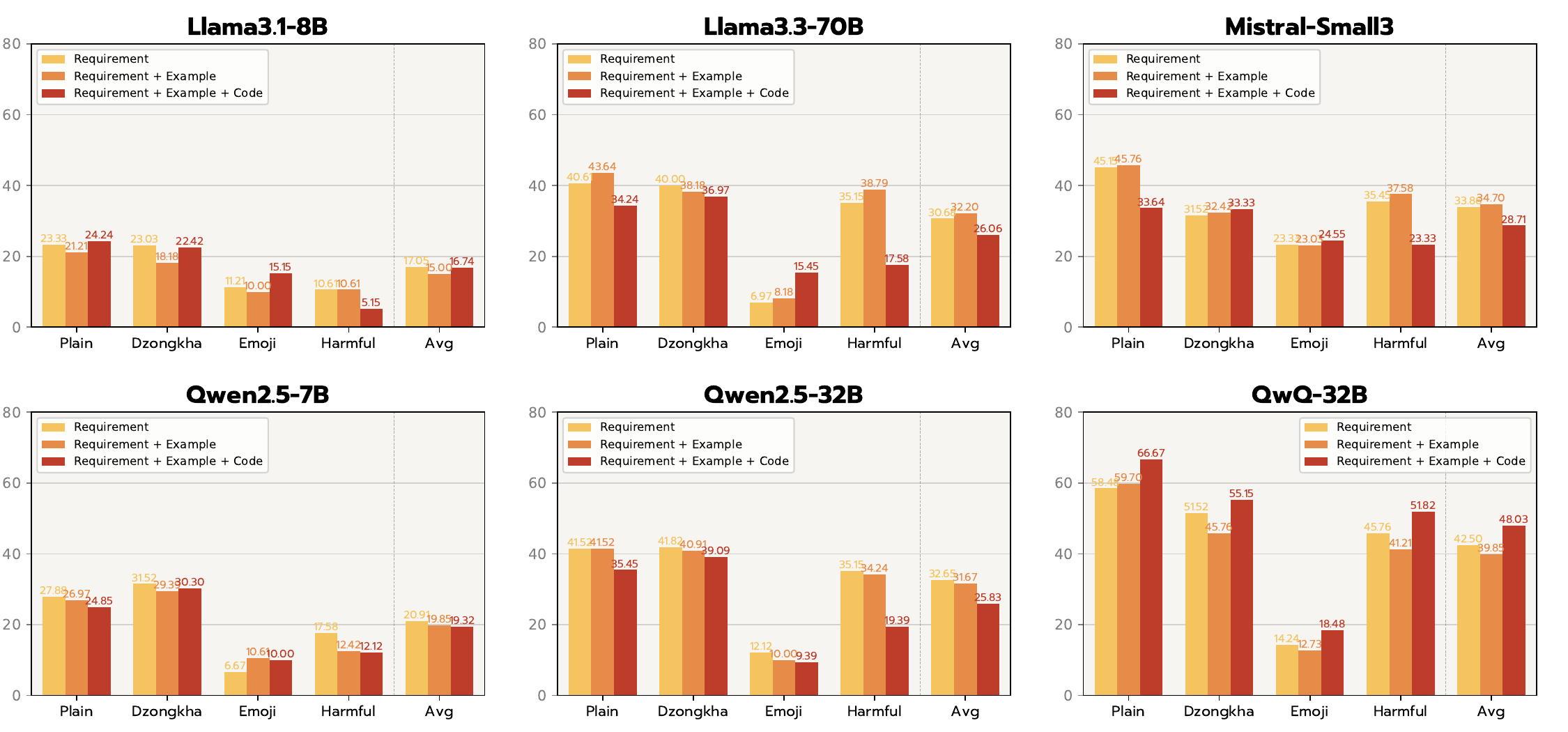}
 \caption{
Performance differences based on the level of rubric details, which are reported using the FA metric}
 \label{fig:rec}
\end{figure*}

\section{Experiments}

In this section, we analyze the performance of various advanced LLMs using MCBench. These experiments aim to demonstrate that our benchmark provides a comprehensive and robust dataset for evaluating LLM performance. We outline our research questions for each experiment and present the insights gained from our findings.

\paragraph{Experimental Settings}
We employ a range of LLMs with different capabilities for extensive analyses. The models we used in our experiments and evaluation prompts are detailed in the Appendix~\ref{app:model_details}. Informed by the \citet{song-etal-2025-good}, which indicates that greedy decoding generally outperforms sampling methods for most evaluated tasks, we set the temperature to 0.0 to ensure deterministic outputs.

\paragraph{RQ1. Do current advanced LLMs possess the capabilities required by our benchmark?} 

Table~\ref{tb:main_results} presents the performance of various LLMs on the MCBench, and Figure \ref{fig:q_fa} demonstrates FA for each metric. As evidenced by the experimental results, our benchmark challenges even the model used for data generation, gpt-4o, which achieves a final accuracy of only 40.98. This demonstrates that our benchmark demands a high level of complex instruction-following capability, underscoring the distinction between "knowing" a concept and "performing" it.

For reasoning models like QwQ, while their FA surpasses that of gpt-4o, their score on format following (FF) is noticeably lower, even lower than 7B-sized models. This suggests that although reasoning can improve task accuracy, it may significantly impair the ability to adhere to the format requirements of instructions. Therefore, enhancing performance on complex instructions, as required by our benchmark, necessitates competencies beyond mere reasoning enhancement.

Additionally, we observe distinct performance differences across categories. Despite the rubric requiring only a straightforward string matching algorithm, these variations indicate that LLMs have significant differences in handling various forms of input.

Table~\ref{tb:sample} displays the expected answers and error cases encountered during evaluation with MCBench. This highlights vulnerabilities in LLMs when dealing with complex instructions requiring mathematical reasoning. First, inadequacies in instruction handling can lead to incorrect outcomes starting from the tokenization phase. Even when tokenization is accurate, insufficient mathematical reasoning skills can result in incorrect calculations. Additionally, while processing complex instructions, formatting instructions may be overlooked. In cases of inadequate long-range consistency, errors such as repeated incorrect phrases in lengthy text outputs may occur. These errors demonstrate that MCBench demands a comprehensive level of competence from LLMs.

\paragraph{RQ2. Does the amount of provided information impact instruction-following performance?}

In constructing a rubric for each metric, we included components such as requirements, examples, and code. We analyze how incorporating these elements affects performance. The experimental results are presented in Figure~\ref{fig:rec}.

The findings indicate that more information does not necessarily lead to better performance. Notably, well-established and self-contained requirement statements alone allow current LLMs to demonstrate a reasonable degree of instruction-following capability, while the addition of code or examples does not consistently enhance performance. In the emoji category, incorporating code generally improved performance, whereas it led to general declines in the harmful category. This suggests that the required attributes vary based on input format, underscoring the effectiveness of our comprehensive benchmark in evaluating diverse scenarios.

Including code in the input proved beneficial for reasoning models but did not typically result in substantial performance gains overall. This might relate to the inherent code comprehension capabilities of LLMs. Models like QwQ, which perform deep input understanding, can achieve significant performance improvements with such detailed inputs, though other models often experienced performance declines.




\paragraph{RQ3. Is there a naming bias associated with using the term "NLP score" when constructing the rubric?}
There may be concerns that inherent familiarity with a specific metric name could negatively impact the abilities we aim to assess in our benchmark. 
To investigate the impact of pre-existing knowledge for each metric, we examine performance variations when using original metric names. The results, as shown in Table~\ref{tb:original}, indicate minimal differences between employing the original naming conventions and the newly introduced "NLP score." This suggests that our benchmark exhibits a certain robustness to intrinsic knowledge, and achieving high performance requires a sufficient capability in instruction understanding.



\definecolor{Gray}{gray}{0.95}
\definecolor{deepgreen}{HTML}{537D5D}

\newcolumntype{a}{>{\columncolor{Gray}}c}

\begin{table}[t]
\centering
\small
\begin{tabular}{l|cca}

\toprule[1.5pt]
\textbf{Model} & \textbf{Original} & \textbf{NLP}& $\Delta$ \\ \midrule[1.5pt]

\textbf{Llama3.1-8B} & 15.45 & \textbf{16.74} & \textcolor{red}{-1.29} \\
\textbf{Qwen2.5-7B} & \textbf{19.55 }& 19.32 & \textcolor{deepgreen}{+0.23} \\
\textbf{Mistral-Small3} & \textbf{29.32} & 28.71 & \textcolor{deepgreen}{+0.61} \\
\textbf{Qwen2.5-32B} & \textbf{26.97} & 25.83 & \textcolor{deepgreen}{+1.14} \\
\textbf{QwQ-32B} & 45.45 & \textbf{48.03} & \textcolor{red}{-2.58} \\
\textbf{Llama3.1-70B} & \textbf{25.45} & 24.55 & \textcolor{deepgreen}{+0.91} \\
\textbf{Llama3.3-70B} & \textbf{26.21} & 26.06 & \textcolor{deepgreen}{+0.15} \\

\bottomrule[1.5pt]
\end{tabular}
\caption{Performance difference between using established original metric names and employing the arbitrary term "NLP score" to describe the rubric
}
\label{tb:original}
\end{table}

\definecolor{metacolorab}{HTML}{F0F0F0}

\begin{table}[t]
\centering
\resizebox{1.0\linewidth}{!}{
\footnotesize
\begin{tabular}{c|c|cll}

\toprule[1.5pt]

\textbf{Category} & \textbf{Measure} & \textbf{Qwen} & \textbf{Qwen-Code} & \textbf{Qwen-Math} \\ \midrule[1.5pt]

\rowcolor{metacolorab} \multicolumn{5}{c}{\textit{Requirements + Example + Code}}\\ \midrule[1.5pt]

\multirow{3}{*}{\textbf{Plain}} & \textbf{FA} & 24.85 & \textbf{28.18} (+3.33) & 20.91 (-3.94) \\
& \textbf{FF} & 63.03 & \textbf{69.09} (+6.06) & 4.85 (-58.18) \\
& \textbf{FD} & \textbf{46.63} & 45.76 (-0.87) & 26.47 (-20.16) \\ \midrule

\multirow{3}{*}{\textbf{Dzongkha}} & \textbf{FA} & \textbf{30.30} & 19.7 (-10.6) & 14.85 (-15.45) \\
& \textbf{FF} & \textbf{65.45} & 60.3 (-5.15) & 9.39 (-56.06) \\
& \textbf{FD} & \textbf{40.81} & 39.49 (-1.32) & 26.09 (-14.72) \\ \midrule

\multirow{3}{*}{\textbf{Emoji}} & \textbf{FA} & 10.00 & \textbf{10.61} (+0.61) & 8.48 (-1.52) \\
& \textbf{FF} & 47.58 & \textbf{63.64} (+16.06) & 3.94 (-43.64) \\
& \textbf{FD} & 30.89 & \textbf{32.18} (+1.29) & 26.21 (-4.68) \\ \midrule

\multirow{3}{*}{\textbf{Harmful}} & \textbf{FA} & \textbf{12.12} & 8.79 (-3.33) & 8.48 (-3.64) \\
& \textbf{FF} & 60.91 & \textbf{67.27} (+6.36) & 3.64 (-57.27) \\
& \textbf{FD} & \textbf{41.37} & 37.01 (-4.36) & 26.09 (-15.28) \\ \midrule

\multirow{3}{*}{\textbf{Avg}} & \textbf{FA} & \textbf{19.32} & 16.82 (-2.50) & 13.18 (-6.14) \\
& \textbf{FF} & 59.24 & \textbf{65.08} (+5.84) & 5.45 (-53.79) \\
& \textbf{FD} & \textbf{39.92} & 38.61 (-1.31) & 26.21 (-13.71) \\

\midrule[1.5pt]
\rowcolor{metacolorab} \multicolumn{5}{c}{\textit{Requirements Only}}\\ \midrule[1.5pt]

\multirow{3}{*}{\textbf{Plain}} & \textbf{FA} & \textbf{27.88} & 24.85 (-3.03) & 16.06 (-11.82) \\ 
& \textbf{FF} & 60.91 & \textbf{84.24} (+23.33) & 7.58 (-53.33)\\
& \textbf{FD} & 39.78 & \textbf{41.8} (+2.02) & 26.15 (-13.63)\\ \midrule

\multirow{3}{*}{\textbf{Dzongkha}} & \textbf{FA} & \textbf{31.52} & 20.61 (-10.91) & 10 (-21.52) \\ 
& \textbf{FF} & 73.33 & \textbf{77.88} (+4.55) & 7.58 (-65.75) \\
& \textbf{FD} & 39.58 & 35.57 (-4.01) & 26.09 (-13.49) \\ \midrule

\multirow{3}{*}{\textbf{Emoji}} & \textbf{FA} & 6.67 & \textbf{6.97} (+0.30) & \textbf{6.97} (+0.30) \\
& \textbf{FF} & 52.73 & \textbf{66.97} (+14.24) & 3.03 (-49.7) \\
& \textbf{FD} & 30.65 & \textbf{30.88} (+0.23) & 26.03 (-4.62)\\ \midrule

\multirow{3}{*}{\textbf{Harmful}} & \textbf{FA} & \textbf{17.58} & 13.03 (-4.55) & 4.24 (-13.34)\\
& \textbf{FF} & 58.48 & \textbf{68.18} (+9.70) & 9.70 (-48.78)\\
& \textbf{FD} & 34.33 & \textbf{34.53} (+0.20) & 26.19 (-8.14)\\ \midrule

\multirow{3}{*}{\textbf{Avg}} & \textbf{FA} & \textbf{20.91} & 16.36 (-4.55) & 9.32 (-11.59)\\
& \textbf{FF} & 61.36 & \textbf{74.32} (+12.96) & 6.97 (-54.39)\\
& \textbf{FD} & \textbf{36.08} & 35.70 (-0.38) & 26.11 (-9.97)\\

\bottomrule[1.5pt]
\end{tabular}
}
\caption{Performance differences between the code-specialized model and the math-specialized model are evaluated. All models report their performance based on Qwen2.5-7B.
}
\label{tb:code_math}
\end{table}


\paragraph{RQ4. Can specializing in Code/Math improve the abilities required by the benchmark?}
We investigate the impact of specialized capabilities on the performance of MCBench. To ensure a fair comparison, we evaluate the performance of the Qwen2.5-7B model released by the Qwen team alongside the concurrently released code and math models. The experimental results are presented in Table~\ref{tb:code_math}.

As the experimental results indicate, although there are some improvements in specific categories for the Code model, overall performance declines are observed in both the code and math specialized models. Notably, the math model shows a significant drop in SF, suggesting considerable decrease in instruction-handling capabilities, even its enhanced mathematical reasoning skills. For the Code model, while SF improves, we witness geneal decreases in FA, possibly due to a diminished ability to analyze and reason through given requirements.

These findings demonstrate that MCBench requires a comprehensive set of capabilities from LLMs. Enhancing a single aspect of performance yields limited benefits; achieving high scores generally necessitates strong performance across all capabilities. This underscores the robustness of our benchmark and its validity as an objective evaluation measure.





\section{Conclusion}
This paper introduces \textbf{MCBench}, a comprehensive and objective benchmark designed to evaluate LLMs. \textbf{MCBench} consists of a step-by-step rubric for executing a string matching metric, candidates for metric calculation, and reference code to objectively assess LLM outputs. By ensuring LLMs strictly follow the provided rubric to compute metrics, we comprehensively evaluated three attributes: complex instruction following, mathematical reasoning, and long-range consistency. We expanded our analysis by introducing four candidate categories, three variants of the rubric components, and three distinct evaluation measures. Our benchmark revealed that even advanced LLMs, such as GPT-4o, achieved only a performance level of 40.98. Enhancing a singular capability, such as code or math specialization, proved minimally effective. Through these analysis, we demonstrated that MCBench is a highly effective objective benchmark for thoroughly assessing LLM capabilities. In future research, we plan to introduce an objective benchmark that incorporates tool implementation.


\section*{Limitation}
While MCBench targets string-matching metrics, other NLP primitives—e.g., probabilistic measures, graph-based scores, or differentiable similarity functions—remain unexplored. Extending the benchmark to these domains will broaden its coverage. 
To isolate the LLM's ability to follow complex instructions, we deliberately excluded external tools such as Python implementations in evaluating LLM. This approach allowed us to elicit the inherent mathematical reasoning capabilities of the LLM.

\section*{Ethics Statement}
We conducted a human inspection of all generated data to ensure there were no ethical issues. The harmful text we used was sourced from the dataset released by \cite{hui2024pleak}, and it does not contain inherent ethical problems. However, if the prompt leads to the generation of harmful responses, it could pose ethical concerns. It is important to note that the purpose of using such data is to assess whether the LLM overreacts to harmful text. We strongly oppose any attempts to solicit responses to these prompts. An AI assistant contributed to the writing of this paper by providing grammar checking and writing support only. The assistant did not contribute to the research content or the development of the study’s topic.

\section*{Acknowledgements}
This work was partly supported by ICT Creative Consilience Program through the Institute of Information \& Communications Technology Planning \& Evaluation(IITP) grant funded by the Korea government(MSIT) (IITP-2025-RS-2020-II201819, 25\%) (RS-2024-00398115, 25\%) (No. RS-2022-II220369, (Part 4) Development of AI Technology to support Expert Decision-making that can Explain the Reasons/Grounds for Judgment Results based on Expert Knowledge, 25\%) and grant funded by Institute of Information \& communications Technology Planning \& Evaluation(IITP) under the Leading Generative AI Human Resources Development(IITP-2025-R2408111, 25\%) grant funded by the Korea government(MSIT).

\bibliography{custom,metrics}


\appendix


\section{Dataset Details}
We report the data statistics of MCBench in Table~\ref{tb:statistics}.
\label{app:stat}

\begin{table}[h]
\centering
\footnotesize
\begin{tabular}{l|ll}

\toprule[1.5pt]
\multicolumn{3}{c}{\textbf{Basic Statistics}} \\ \midrule
\textbf{\# Rubric} & \multicolumn{2}{c}{33}\\
\textbf{\# Candidates} & \multicolumn{2}{c}{40} \\
\textbf{\# Test Instances} & \multicolumn{2}{c}{1,320} \\
\textbf{Avg/Min/Max \# Steps} & \multicolumn{2}{c}{4.18 / 3 / 7} \\ \midrule
\multicolumn{3}{c}{\textbf{Avg \# character - Rubric}} \\ \midrule
\textbf{Requirement} & \multicolumn{2}{c}{2259.3} \\
\textbf{Requirement + Example} & \multicolumn{2}{c}{3096.2} \\
\textbf{Requirement + Example + Code} & \multicolumn{2}{c}{5145.0} \\ \midrule
\multicolumn{3}{c}{\textbf{Avg \# character - Candidate}} \\ \midrule
\textbf{Plain} & \multicolumn{2}{c}{24.3} \\
\textbf{Dzongkha} & \multicolumn{2}{c}{31.2} \\
\textbf{Emoji} & \multicolumn{2}{c}{16.4} \\
\textbf{Harmful} & \multicolumn{2}{c}{46.3} \\

\bottomrule[1.5pt]
\end{tabular}
\caption{Data Statistics.
}
\label{tb:statistics}
\end{table}

To determine candidates, we organized data such that character lengths were similar across categories to minimize difficulty differences. We avoided constructing samples with excessively long character lengths, as processing such candidates would also take longer.

The character length of plain candidates served as the baseline. To generate these plain candidates, we prompted GPT-4o with the following query: \textit{"Recommend several examples that are appropriate to measure..."}. We assumed the character lengths of these generated candidates as standard. We ensured that the data for Dzongkha, Emoji, and Harmful categories did not significantly deviate from the length of these plain candidates.

For Dzongkha and Emoji, we used prompts such as, \textit{"Recommend syntactically similar sentence pairs written in \{Emoji, Dzongkha\}"} to generate data. Initially, candidates were generated using these prompts, then the authors select and adjust the data to meet length criteria. We utilized GPT-4o throughout this data generation process.

For the Harmful data category, we adopted the dataset proposed by \cite{hui2024pleak} to construct our benchmark. We focused on the shortest data instances to consider character length. However, this data tends to be twice as long as plain candidates, which might introduce inherent difficulty.

\section{Evaluation Criteria for Human Evaluation}
\label{app:eval}

For the \textbf{Requirement}, we focus on verifying that the descriptions do not contradict the reference documentation. We design our requirements by combining purposefully generated content from three models. If a description is inaccurate or requires additional explanations for a step, we regenerate the requirement for that step to ensure quality. Crucially, we confirm the completeness by verifying that the requirement is self-contained, allowing the metric to be computed directly from the description alone.

For \textbf{Example} and \textbf{Code}, we focus on verifying that they do not contradict the requirements. In verifying \textbf{Example}, we ensure they accurately illustrate each step with concise execution instances. Similarly, for \textbf{Code}, we check that each step is appropriately addressed. Specifically, the code undergoes actual Python implementation, and if any execution errors arise, the authors directly correct them to ensure functionality.

\section{Model Details}
\label{app:model_details}
Table~\ref{tb:model_details} shows several LLMs we employed for our experiments and data construction. By conducting experiments with LLMs of varying performance levels, we aim to verify the robustness of our benchmark.

\begin{table}[h]
\centering

\resizebox{0.95\linewidth}{!}{
\begin{tabular}{l|c}

\toprule[1.5pt]
\makecell[c]{\textbf{Model Name}}& \makecell[c]{{\# Params}} \\ \midrule[1.5pt]

\textbf{Llama3.1-8B} \cite{grattafiori2024llama} & \multirow{2}{*}{8.03B} \\
: \texttt{meta-llama/Meta-Llama-3.1-8B-Instruct} & \\ \midrule

\textbf{Llama3.1-70B} \cite{grattafiori2024llama} & \multirow{2}{*}{70.6B} \\
: \texttt{meta-llama/Llama-3.1-70B-Instruct} & \\ \midrule

\textbf{Llama3.3-70B} \cite{grattafiori2024llama} & \multirow{2}{*}{70.6B} \\
: \texttt{meta-llama/Llama-3.3-70B-Instruct} & \\ \midrule

\textbf{Mistral-Small3} \cite{mistral3}  & \multirow{2}{*}{23.6B} \\
: \texttt{mistralai/Mistral-Small-24B-Instruct-2501} & \\ \midrule

\textbf{Qwen2.5-7B} \cite{qwen2.5} & \multirow{2}{*}{7.62B} \\
: \texttt{Qwen/Qwen2.5-7B-Instruct} & \\ \midrule

\textbf{Qwen2.5-32B} \cite{qwen2.5} & \multirow{2}{*}{32.8B} \\
: \texttt{Qwen/Qwen2.5-32B-Instruct} & \\ \midrule

\textbf{QwQ-32B} \cite{qwq32b} & \multirow{2}{*}{32.8B} \\
: \texttt{Qwen/QwQ-32B} & \\ \midrule

\textbf{Llama3.3-70B(R1)} \cite{guo2025deepseek} & \multirow{2}{*}{70.6B} \\
: \texttt{deepseek-ai/DeepSeek-R1-Distill-Llama-70B} & \\ \midrule

\textbf{Qwen2.5-32B(R1)} \cite{guo2025deepseek} & \multirow{2}{*}{32.8B} \\
: \texttt{deepseek-ai/DeepSeek-R1-Distill-Qwen-32B} & \\ \midrule

\textbf{Qwen2.5-Coder-7B} \cite{qwencode} & \multirow{2}{*}{7.62B} \\
: \texttt{Qwen/Qwen2.5-Coder-7B-Instruct} & \\ \midrule

\textbf{Qwen2.5-Math-7B} \cite{qwenmath} & \multirow{2}{*}{7.62B} \\
: \texttt{Qwen/Qwen2.5-Math-7B-Instruct} & \\ \midrule

\textbf{gpt-4o-mini} \cite{hurst2024gpt} & \multirow{2}{*}{-} \\
: \texttt{gpt-4o-mini-2024-07-18} & \\ \midrule

\textbf{gpt-4o} \cite{hurst2024gpt} & \multirow{2}{*}{-} \\
: \texttt{gpt-4o-2024-08-06} & \\\midrule

\textbf{o1} \cite{jaech2024o1} & \multirow{2}{*}{-} \\
: \texttt{o1-2024-12-17} & \\\midrule

\textbf{o3-mini} \cite{o3mini} & \multirow{2}{*}{-} \\
: \texttt{o3-mini-2025-01-31} & \\

\bottomrule[1.5pt]

\end{tabular}
}\caption{Model Details. We deployed OPENAI API call for experiments with GPT-4o-mini and GPT-4o, and \texttt{huggingface} \cite{wolf-etal-2020-transformers} for eliciting model weights for other publicly-available LLMs.
}\label{tb:model_details}
\end{table}

\section{Evaluation Details}
To evaluate the performance of MCBench, we provided each model with the prompts as shown in Table~\ref{tb:prompt_eval}. In this prompt, \textbf{Statement A} and \textbf{Statement B} refer to the two sentence pairs that comprise the \textbf{Candidates}.
\definecolor{metacolorab}{HTML}{F0F0F0}

\begin{table}[h]
\centering
\resizebox{1.0\linewidth}{!}{
\begin{tabular}{l}

\toprule[1.5pt]

\rowcolor{metacolorab} \textbf{System Prompt} \\ \midrule[1.5pt]

\makecell[l]{You are an expert in computer science and a highly capable, \\ responsive assistant trained to assist with a broad range of tasks. \\
I have derived a new metric, \textbf{\{\textit{NLP score}\}}. \\
\textbf{\{\textit{NLP score}\}} is an automatic evaluation metric used for \\ 
comparing the similarity between a hypothesis and reference text. \\
I will show you the detailed process to calculate this metric. \\
Your task is to derive the \textbf{\{\textit{NLP score}\}} for the given input values  \\
by referring to the provided process. \\
At each intermediate step, print the corresponding result. \\
Before generating the final answer, briefly explain your reasoning \\ behind it. You must find your final answer before your response \\reaches 1000 lines.
Finally, compile all the results and output your \\ 
final verdict by strictly following this format: \\\\
\textbf{\#\#\# Final Results \#\#\#} \\
\text{[Step1]} : ... \\
\text{[StepN]} : ... \\
\text{[Final]} : ...} \\ \midrule[1.5pt]

\rowcolor{metacolorab} \textbf{User Query} \\ \midrule[1.5pt]
\textbf{\{\textit{Rubric}\}} \\\\

Tell me the \textbf{\{\textit{NLP score}\}} of "a" with a reference text "b", \\
following the previous rubric: \\
a = \textbf{\{\textit{Statement A}\}} \\
b = \textbf{\{\textit{Statement B}\}} \\
\bottomrule[1.5pt]

\end{tabular}
}
\caption{Prompt for evaluating LLMs on MCBench} \label{tb:prompt_eval}
\end{table}

\section{Data Construction Details}
The prompt we used for data construction is shown in Table~\ref{tb:prompt_data}.
\definecolor{metacolorab}{HTML}{F0F0F0}

\begin{table}[h]
\centering
\resizebox{1.0\linewidth}{!}{
\begin{tabular}{l}

\toprule[1.5pt]

\rowcolor{metacolorab} \textbf{System Prompt} \\ \midrule[1.5pt]
You are an expert in computer science and a highly capable, \\ responsive assistant trained to assist with a broad range of tasks.  \\
Your main goal is to provide helpful, relevant, and accurate  \\
information to users. 
Your responses should be clear and adaptable \\
to different levels of user expertise. \\
Be concise but detail-oriented. \\
\midrule[1.5pt]

\rowcolor{metacolorab} \textbf{User Query} \\ \midrule[1.5pt]
I want to calculate \textbf{\{\textit{Metric Name}\}} between statement A and B. \\ \\
Explain necessary detailed procedures step-by-step. \\
Please explain the processes required at each step in a clear and \\
detailed manner.  \\
Please separate each step with “\#\#\# Step N”.   \\ \\
Your explanation for each step must include the following  \\
three parts: "Requirement", "Example", and "Code".  \\
Detailed instructions are as follows: \\ \\
- Requirement Part  \\
You must explain detailed procedures required for each step.  \\
Descriptions should be self-contained that anyone can  \\
re-implement it without referring other documents. \\
Your explanation should be concise but contain sufficient details. \\
You must make a response in a definitive way. \\
If necessary, separate subprocesses for each step. \\
Requirement part should not contain any code or examples.  \\
If formulas are necessary, output them in LaTeX format.  \\
Begin your descriptions with "[Requirement]". \\ \\
- Example Part \\
For each step, you must include a simple implementation example.  \\
Provided example should be aligned with Requirement. \\
Begin your descriptions with "\text{[Example]}". \\
 \\
- Code Part \\
You must provide Python code for each step in a way that it can  \\
be executed without any packages, implemented as functions.  \\
Name of each function should be "stepN" that corresponds \\
to each step. \\
After providing your function code, give me an example usage of it. \\
Provided example should be aligned with Requirement. \\
Begin your descriptions with "\text{[Python Code]}". \\
\bottomrule[1.5pt]

\end{tabular}
}
\caption{Prompt for constructing MCBench. Note that every data generated with this prompt were later refined through human inspection.} \label{tb:prompt_data}
\end{table}

\section{Justification for Selecting String-Match Metrics}
Our benchmark primarily focuses on string-matching metrics for several reasons:
\begin{itemize}
    \item It does not require contextual understanding of "candidate". This approach isolates the ability to "understand and execute instructions" from the ability to "process given inputs." This allows us to analyze the limitations of LLMs' inherent input understanding by evaluating performance variations based on different type of candidates. Defining metrics enables constructing parallel reference code and objective evaluation of scores. Clearly defined outputs allow for objective assessments, which is a significant advantage of our benchmark.
    \item It effectively evaluates long-range consistency. This "long-range"includes not only inputs composed of multi-step instructions but also extended solutions required to derive the final answer. The benchmark is designed to assess LLMs’ ability to maintain consistency, follow instructions, and perform numerical reasoning for accurate outcomes with long input-output interactions.
    \item It comprehensively evaluates mathematical reasoning and instruction following. Unlike benchmarks like AIME or MATH-500, where the LLM must "reason" to find a path for solutions, this benchmark requires intact adherence to given instructions while also demonstrating mathematical skills. While data sorting and financial calculations could be considered under this purpose, they seem beyond our benchmark's defined scope. A broader definition would be necessary to encompass these tasks, but we view that presenting such an objective and experimental scope is slightly beyond our current focus.
\end{itemize}

We find it impractical to consider all code-implementable measures, as indiscriminate inclusion could obscure the benchmark's purpose. While future research may pursue a more comprehensive definition, we believe focusing on our initial scope is preferable. Therefore, we clearly define our objectives and systematically present findings within that scope.



\end{document}